\documentclass[journal]{IEEEtran}

\usepackage{cite}
\usepackage{amsmath,amssymb,amsfonts}
\usepackage{graphicx}
\usepackage{algorithm,algorithmic}
\usepackage{booktabs}
\usepackage{tabularx}
\usepackage{array}
\usepackage{multirow}
\usepackage{textcomp}
\usepackage{hyperref}
\hypersetup{hidelinks}
\usepackage{threeparttable}

\def\BibTeX{{\rm B\kern-.05em{\sc i\kern-.025em b}\kern-.08em
    T\kern-.1667em\lower.7ex\hbox{E}\kern-.125emX}}

\newcommand{\mypm}{$\pm$}

\begin{document}

\title{EEG-VID: Task-Guided Latent Predictive Pretraining for EEG Decoding and Assistive Target Selection}

\author{%
Guanzhong Sun, Junyi Ma, Yuxuan Wu,Wei Tang, and Yanzi Miao%
\thanks{Guanzhong Sun and Yanzi Miao are with the School of Information and
Control Engineering, China University of Mining and Technology, Xuzhou, China.}%
\thanks{Junyi Ma is with the School of Automation and Intelligent Sensing,
Shanghai Jiao Tong University, Shanghai, China.}%
\thanks{Yuxuan Wu is with the School of Automation and Intelligent Sensing,
Shanghai Jiao Tong University, Shanghai, China, and also with the Shanghai
Innovation Institute, Shanghai, China.}%
\thanks{Wei Tang is with the School of Mechanical and Electrical Engineering,
China University of Mining and Technology, Xuzhou, China.}%
}

\maketitle

\begin{abstract}
\textit{Objective:} We investigate whether task-guided latent predictive pretraining improves EEG decoding under session and subject shifts and whether a four-electrode spatial posterior can support assistive target selection.

\textit{Methods:} EEG-VID predicts future latent EEG states from recent history using an exponential-moving-average (EMA) target encoder and weak task guidance, and then fine-tunes the pretrained model for supervised decoding. We apply the same Stage~1 objective to five EEG backbones and replace the proposed predictor with recurrent and Transformer alternatives to test transferability and predictor dependence.

\textit{Results:} On the 48-region cross-day VIG-48 task, EEG-VID reaches
6.52\% Top-1 and 30.50\% Top-5 accuracy. Across VIG-48 and BCI Competition IV-2a/IV-2b, Stage~1 improves mean accuracy in 41 of 42 matched backbone--dataset--protocol comparisons, including all 12 leave-one-subject-out (LOSO) settings, with a maximum gain of 16.22 percentage points. Weak task
guidance improves over latent-only pretraining in all 18 pooled
model--dataset comparisons. Alternative predictors outperform direct supervision in 48 of 54 comparisons, while the proposed predictor performs best in 45 of 54 matched comparisons. In a separate six-participant offline robot-scene study, candidate-constrained target selection reaches 40.24\% versus a 25\% chance level after subject-specific calibration.
Because gaze is unconstrained in VIG-48, its spatial result is interpreted at
the wearable-device level rather than as source-specific cortical decoding.

\textit{Conclusion:} Weak task guidance improves the consistency of latent
predictive pretraining across EEG decoders and distribution shifts, while the
resulting spatial posterior supports above-chance scene-constrained target
selection.

\textit{Significance:} Task-guided latent prediction provides a transferable pretraining strategy for EEG decoding under session and subject shifts and connects low-channel EEG spatial posteriors with scene constraints for assistive target selection.
\end{abstract}


\section{Introduction}
\label{sec:introduction}

\IEEEPARstart{R}{eliable} target selection is a basic requirement for
assistive manipulation. Before a robot can reach or grasp an object, it must
infer which target the user intends to select. EEG provides a non-invasive control signal for users with limited speech or manual input, but
practical decoding remains difficult because the signal is noisy and varies
across sessions and users. Established convolutional and transformer-based EEG
decoders can learn effective discriminative features
\cite{Schirrmeister_2017,lawhern2018eegnet,song2022eeg,sun2023survey}, yet
those features may shift with electrode placement, contact impedance, fatigue,
and other non-stationary factors.

\begin{figure*}[!t]
    \centering
    \includegraphics[width=0.75\textwidth]{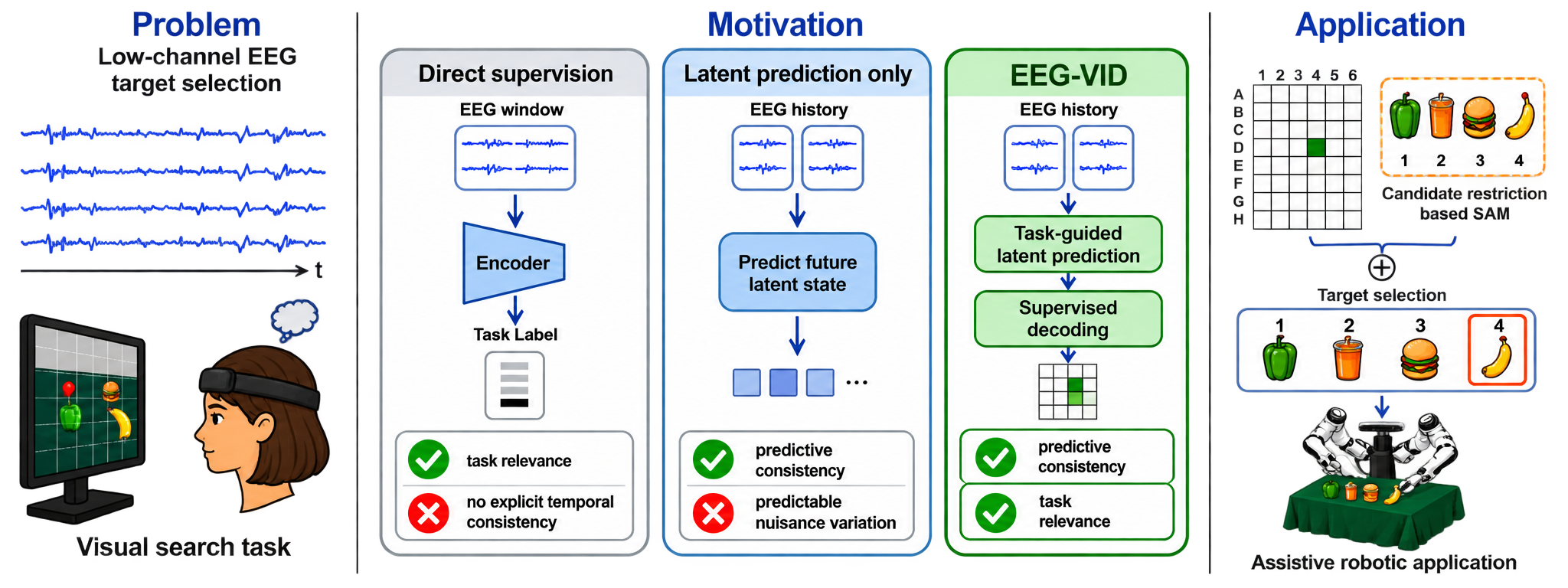}
    \caption{\textbf{Motivation and paradigm comparison of EEG-VID.}
    EEG-VID combines temporal predictive consistency with weak task guidance,
    and restricts the decoded grid-region posterior to scene-derived candidates
    for assistive target selection.}
    \label{fig:motivation}
\end{figure*}

Supervised EEG decoders optimize task discrimination but do not explicitly model cross-window predictive structure. Latent prediction can capture temporal dynamics, yet it may also preserve predictable task-irrelevant components such as slow drift, ocular activity, or visually evoked responses. EEG-VID therefore adds weak task guidance to future latent-state prediction so that pretraining emphasizes predictable components relevant to decoding
(Fig.~\ref{fig:motivation}).

Based on this observation, we develop EEG-VID with task-guided latent
predictive pretraining (Stage~1) followed by supervised EEG decoding
(Stage~2). We focus on whether this pretraining objective transfers across different EEG decoders under session and subject shifts, rather than whether it benefits EEG-VID alone.

We evaluate this question on the in-house VIG-48 cross-day dataset and BCI
Competition IV-2a/IV-2b
\cite{brunner2008bci2a,leeb2008bci2b,tangermann2012review} using pooled
cross-session, within-subject, and leave-one-subject-out protocols. Stage~1 is
applied to five existing EEG backbones
\cite{Schirrmeister_2017,lawhern2018eegnet,song2022eeg,
song2024decoding,guo2025neuro}, and the proposed predictor is replaced by
recurrent and Transformer alternatives to further test whether the observed
benefit depends on a particular predictor architecture.

A separate six-participant offline robot-scene study examines the assistive
relevance of the learned spatial posterior under scene-constrained target
selection. Across all evaluations, Stage~1 improves mean accuracy in 41 of 42 matched backbone--dataset--protocol comparisons, including all 12 LOSO settings. Alternative predictors outperform direct supervision in 48 of 54 comparisons. Robot-scene selection reaches
40.24\% versus a 25\% chance level after subject-specific calibration.
Because gaze is unconstrained in VIG-48, we interpret its spatial decoding
result at the wearable-device level rather than as source-specific cortical
decoding.

The main contributions are:
\begin{itemize}

\item We propose a task-guided latent predictive pretraining objective that combines future-state prediction with weak task supervision to learn predictable EEG representations that remain useful for downstream decoding.

\item We formulate Stage~1 as a backbone-agnostic pretraining objective and evaluate its transfer through matched comparisons across different EEG decoders, predictor architectures, and session and subject shift protocols.

\item We evaluate the proposed strategy in a low-channel cross-day visual decoding setting and further test its use for scene-constrained target selection in a six-participant offline robot-scene study.

\end{itemize}

\section{Related Work}

\subsection{EEG-based intention decoding}
EEG has long been used to decode user intentions and cognitive states in BCIs and rehabilitation systems. Classical paradigms include motor imagery
\cite{lawhern2018eegnet,Schirrmeister_2017}, steady-state visual evoked
potentials \cite{Waytowich_2018}, P300 event-related potentials
\cite{farwell1988talking}, and visual stimulus classification
\cite{song2024decoding,guo2025neuro,yao2025visualattention}. Recent studies extend EEG-based visual
interfaces to large-scale natural-image and object decoding
\cite{gifford2022large,song2024decoding,wu2025bridging},
brain-supervised image editing \cite{davis2022brain}, and 3D visual
perception \cite{guo2025neuro}. Beyond decoding alone, assistive BCI systems
have combined neural intent with autonomous or shared robotic control
\cite{millan2010combining,iturrate2009wheelchair,carlson2013wheelchairs,
pmlr-v229-zhang23f,xu2019shared,zhou2023shared}. Most of these systems are trained with discriminative objectives on fixed or aggregated observation windows. Our work instead learns predictive structure in latent EEG dynamics before supervised decoding.

\subsection{EEG representation learning and adaptation}
EEG representation learning is challenged by low signal-to-noise ratio,
non-stationarity, and substantial subject/session variability \cite{zhou2023interpretable}. These
distribution shifts have motivated transfer, alignment, and
subject-independent learning for cross-session and cross-subject EEG decoding
\cite{jayaram2016transfer,zanini2018transfer,rodrigues2019riemannian,
wu2022transfer,He2018TransferLF,flores2024active,she2023wasserstein,
sartipi2024subjectindependent,zhong2024tactile,zhou2023spatial,chen2025uda}.
Existing approaches further include signal preprocessing and channel
selection \cite{sun2023survey}, spatiotemporal neural architectures
\cite{Schirrmeister_2017,lawhern2018eegnet,song2022eeg,li2021tsseff},
and lightweight or subject-specific calibration.

Self-supervised and pretrained EEG representation learning spans contrastive,
temporal-context, and other pretext objectives
\cite{mohsenvand2020contrastive,banville2021uncovering,kostas2021bendr,
rafiei2024ssl}, masked or predictive representation learning
\cite{foumani2024eeg2rep,fu2024gmaeeg}, and larger cross-dataset pretrained
models \cite{yang2023biot,jiang2024labram,wang2024eegpt,wang2025cbramod}.
Purely self-supervised predictive objectives do not explicitly prioritize predictable EEG components that are informative for a specific downstream task.

\subsection{Latent predictive modeling}
Predictive representation learning models future or missing information in
representation space rather than reconstructing raw observations. Contrastive
Predictive Coding learns representations by predicting future latent states
\cite{oord2018cpc}, while data2vec predicts contextualized latent targets
\cite{baevski2022data2vec}. Target-encoder methods such as BYOL and I-JEPA
further demonstrate predictive learning directly in representation space
\cite{grill2020bootstrap,assran2023self}. For EEG, latent prediction avoids raw-signal reconstruction, which may require the model to represent task-irrelevant signal variation. The key challenge is to retain predictable components that are informative for the downstream task. We study whether weak task guidance can retain task-relevant structure within predictive EEG representations across backbones and subject shifts.

\section{Methods}
\label{sec:method}

\begin{figure*}[!t]
\centerline{\includegraphics[width=\textwidth]{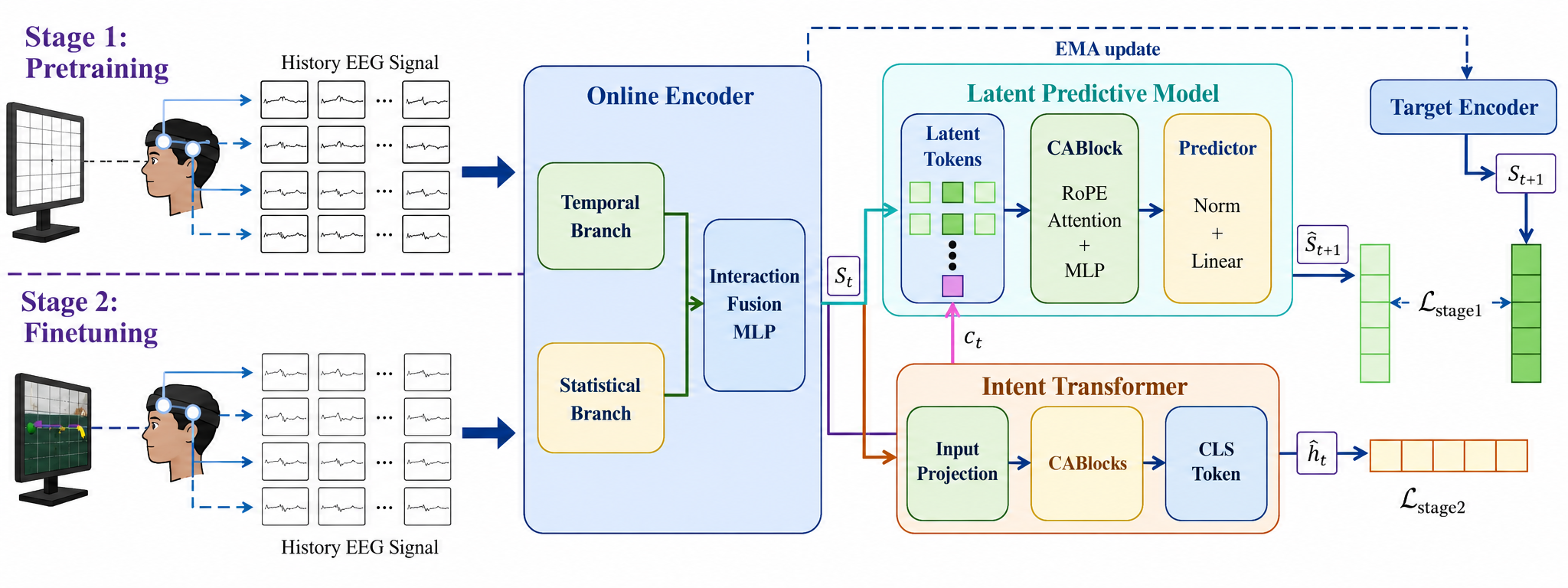}}
\caption{\textbf{Overall architecture of EEG-VID.} Stage~1 learns latent
EEG-state transition dynamics by predicting future EEG representations from
historical EEG windows, supervised by an EMA target encoder and guided by a weak auxiliary classification term. Stage~2 transfers the learned dynamic
representation to supervised visual intention region decoding.}
\label{fig:architecture}
\end{figure*}

This section presents EEG-VID, a two-stage framework for region-level visual
intention decoding (Fig.~\ref{fig:architecture}). EEG-VID first learns latent
transitions from historical to future EEG representations through latent predictive
pretraining, and then transfers the learned dynamic representations to
supervised region decoding. The key idea is that future latent-state prediction provides a temporal
consistency constraint, while weak task guidance encourages the learned
predictable structure to remain relevant to downstream decoding.

\subsection{Problem Formulation}
For each visual target selection trial, the user observes a visual scene
divided into an $N\times M$ grid, corresponding to $NM$ candidate regions.
Given a sequence of historical EEG windows
\[
\mathbf{X}_{t}=\{\mathbf{x}_{t-K+1},\ldots,\mathbf{x}_{t}\},
\]
where $\mathbf{x}_{t}\in\mathbb{R}^{L\times C}$ denotes the $t$-th EEG window
and $K$, $L$, $C$ denote the number of historical windows, the window length
and the number of EEG channels, the goal is to predict the region label
$h_t\in\{0,\ldots,NM-1\}$ of the user's intended target.

Rather than decoding each EEG window independently, EEG-VID models temporal transitions between historical and future latent EEG states. During training, the model learns to predict future latent EEG representations from historical windows. At inference, only $X_t$ is required.

\subsection{Multi-scale Temporal-Statistical Hybrid EEG Encoder}
\begin{figure}[!t]
\centerline{\includegraphics[width=\columnwidth]{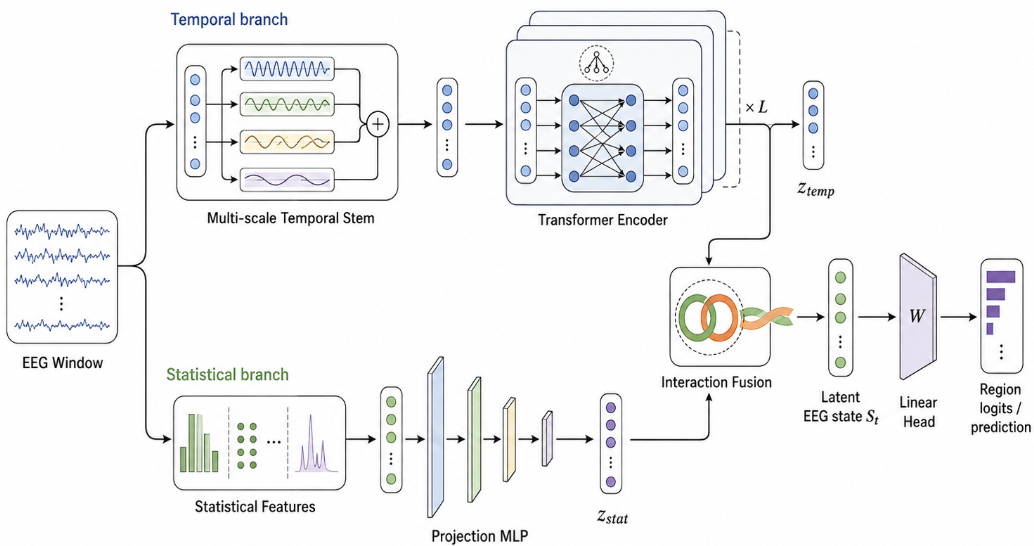}}
\caption{\textbf{Architecture of the multi-scale temporal-statistical hybrid
EEG encoder.} A temporal branch extracts multi-scale temporal patterns, while a statistical branch extracts stable statistical descriptors. The two branches are fused through interaction features and projected into the EEG latent representation.}
\label{fig:encoder_detail}
\end{figure}

The Online Encoder $E_{\theta}$ maps each EEG window to a latent EEG-state
representation. To combine learned temporal patterns with stable signal descriptors, we use a two-branch encoder (Fig.~\ref{fig:encoder_detail}). The temporal branch models multi-scale dynamics, while the statistical branch summarizes additional signal statistics.

\subsubsection{Temporal branch}
A multi-scale temporal stem applies parallel one-dimensional convolutions with
kernel sizes $\{3,7,11,15\}$,
\begin{equation}
\mathbf{u}^{(k)}_{\tau}=\mathrm{Conv1D}_{k}(\mathbf{x}_{\tau}),
\qquad k\in\mathcal{K},
\end{equation}
whose outputs are concatenated along the channel dimension, normalized,
activated by GELU and projected to a common embedding dimension by a
$1\times1$ convolution,
\begin{equation}
\mathbf{U}_{\tau}=\mathrm{FrontProj}\!\left(\mathrm{Concat}_{k\in\mathcal{K}}
\mathbf{u}^{(k)}_{\tau}\right),
\end{equation}
followed by BatchNorm, GELU and a channel squeeze-and-excitation module. Two
dilated residual blocks (dilation $1$ and $2$) enlarge the receptive field,
learnable positional embeddings are added, and a Transformer encoder captures
long-range dependencies within the window,
\begin{equation}
\mathbf{H}_{\tau}=\mathrm{Transformer}\!\left(
\mathrm{ResBlock}(\mathbf{U}_{\tau})+\mathbf{P}\right).
\end{equation}
LayerNorm and attention pooling then yield the window-level temporal
representation
$\mathbf{z}^{\mathrm{temp}}_{\tau}=
\mathrm{AttnPool}(\mathrm{LayerNorm}(\mathbf{H}_{\tau}))$.

\subsubsection{Statistical branch}
For each window, we compute six channel-wise temporal descriptors---mean,
standard deviation, RMS, absolute mean, first-difference standard deviation,
and line length---together with pairwise channel correlations and binned
log-power spectral features. Their concatenation
$\mathbf{r}_{\tau}=[\mathbf{r}^{\mathrm{time}}_{\tau}\|
\mathbf{r}^{\mathrm{corr}}_{\tau}\|\mathbf{r}^{\mathrm{freq}}_{\tau}]$
is mapped by a projection MLP to the statistical representation
$\mathbf{z}^{\mathrm{stat}}_{\tau}=P_{\eta}(\mathbf{r}_{\tau})$.

\subsubsection{Interaction fusion}
The two branches are fused through an explicit interaction feature
\begin{equation}
\mathbf{q}_{\tau}=\left[
\mathbf{z}^{\mathrm{temp}}_{\tau}\;\|\;
\mathbf{z}^{\mathrm{stat}}_{\tau}\;\|\;
\mathbf{z}^{\mathrm{temp}}_{\tau}\odot\mathbf{z}^{\mathrm{stat}}_{\tau}\;\|\;
\left|\mathbf{z}^{\mathrm{temp}}_{\tau}-\mathbf{z}^{\mathrm{stat}}_{\tau}\right|
\right],
\end{equation}
where $\|$ is concatenation and $\odot$ element-wise multiplication. The product term captures interactions between the two branches, while the absolute-difference term captures differences between their features. A fusion MLP
(LayerNorm--Linear--GELU--Dropout--Linear) produces the window-level latent
EEG state
\begin{equation}
\mathbf{S}_{\tau}=F_{\theta}(\mathbf{q}_{\tau}),\qquad
\mathbf{S}_{\tau}\in\mathbb{R}^{d},
\end{equation}
and we write the complete mapping as
$\mathbf{S}_{\tau}=E_{\theta}(\mathbf{x}_{\tau})$. Applying $E_{\theta}$ to
each window with shared parameters gives the historical latent sequence
\begin{equation}
\mathbf{S}_{t-K+1:t}=E_{\theta}(\mathbf{X}_{t})\in\mathbb{R}^{K\times d}.
\end{equation}

\subsection{Task-Guided Latent Predictive Pretraining (Stage 1)}
Given $\mathbf{S}_{t-K+1:t}$, an intention modeling branch aggregates the
historical latent tokens and outputs an intention representation
$\mathbf{c}_t$ together with auxiliary region logits
$\mathbf{o}^{\mathrm{aux}}_t$:
\begin{equation}
(\mathbf{c}_t,\mathbf{o}^{\mathrm{aux}}_t)=
G_{\psi}(\mathbf{S}_{t-K+1:t}),\qquad
\mathbf{o}^{\mathrm{aux}}_t\in\mathbb{R}^{NM}.
\end{equation}
$G_{\psi}$ is an Intent Transformer operating only on the $K$
historical latent tokens. A learnable summary token produces
$\mathbf{c}_t$, and masked self-attention prevents access to the future EEG
window used as the prediction target. The summary-token mask is defined so
that it can aggregate all historical tokens while remaining independent of
$\mathbf{x}_{t+1}$.

The latent predictor is conditioned on both the history and the
intention representation,
\begin{equation}
(\hat{\mathbf{S}}_{t+1},\mathbf{w}_t)=
W_{\phi}(\mathbf{S}_{t-K+1:t},\mathbf{c}_t),
\end{equation}
where $\hat{\mathbf{S}}_{t+1}$ is the predicted future latent EEG state and
$\mathbf{w}_t$ is the EEG-state dynamics representation. Structurally,
$W_{\phi}$ contains a latent-token projection module, a stack of predictor
blocks, and a lightweight predictor head. Each predictor block combines
rotary positional embeddings (RoPE), attention, and an MLP.

Following the target-encoder principle used in predictive representation
learning \cite{grill2020bootstrap,assran2023self}, the prediction target is
produced by an EMA target encoder $E_{\bar{\theta}}$ that shares the
architecture of $E_{\theta}$ but is updated without gradients,
\begin{equation}
\bar{\theta}\leftarrow\lambda_{\mathrm{ema}}\bar{\theta}
+(1-\lambda_{\mathrm{ema}})\theta,\qquad
\mathbf{S}_{t+1}=E_{\bar\theta}(\mathbf{x}_{t+1}),
\end{equation}
where the target-encoder EMA decay is fixed to
$\lambda_{\mathrm{ema}}=0.995$. A stop-gradient $\mathrm{sg}(\cdot)$ is
applied to $\mathbf{S}_{t+1}$. Together with the EMA target encoder, this stabilizes the prediction target and helps prevent trivial co-adaptation between the online encoder and the target representation.

We separate the magnitude and directional components of future-state
prediction as
\begin{equation}
\mathcal{L}_{\mathrm{latent}}=
\left\|\hat{\mathbf{S}}_{t+1}-\mathrm{sg}(\mathbf{S}_{t+1})\right\|_2^2,
\end{equation}

\begin{equation}
\mathcal{L}_{\mathrm{cos}}=
1-\cos\left(\hat{\mathbf{S}}_{t+1},
\mathrm{sg}(\mathbf{S}_{t+1})\right).
\end{equation}

\subsubsection*{Weak task guidance}
Temporal prediction alone does not determine which predictable EEG components
are useful for downstream decoding because slowly varying task-irrelevant activity can also be predictable. We therefore introduce a weak auxiliary
intention-classification term

\begin{equation}
\mathcal{L}_{\mathrm{intent}}=
\mathrm{CE}(\mathbf{o}^{\mathrm{aux}}_t,h_t).
\end{equation}
The complete Stage~1 objective is
\begin{equation}
\mathcal{L}_{S1}=
\lambda_{\mathrm{latent}}\mathcal{L}_{\mathrm{latent}}
+\lambda_{\mathrm{cos}}\mathcal{L}_{\mathrm{cos}}
+\lambda_{\mathrm{cls}}^{S1}\mathcal{L}_{\mathrm{intent}}.
\label{eq:stage1}
\end{equation}
Unless otherwise stated, we use
$\lambda_{\mathrm{latent}}=1.0$,
$\lambda_{\mathrm{cos}}=0.1$,
$\lambda_{\mathrm{cls}}^{S1}=0.1$, and
$\lambda_{\mathrm{ema}}=0.995$.
The small classification weight allows this term to guide the learned
representation without replacing the latent prediction objective. Accordingly,
Section~\ref{sec:stage1_ablation} compares the default with
$\lambda_{\mathrm{cls}}^{S1}\in\{0,0.2,0.3\}$ while keeping the other
Stage~1 coefficients fixed. This sweep tests the balance between prediction and task supervision over these four settings. It does not show that 0.1 is globally optimal.

Because Stage~1 uses the training-split labels through
$\mathcal{L}_{\mathrm{intent}}$, the procedure is a weakly supervised
pretraining stage rather than a self-supervised one.

\subsection{Supervised Visual Intention Region Decoding (Stage 2)}
Stage~2 transfers the online encoder, the intention branch and the latent predictor
to supervised decoding. Instead of relying on the current window alone, the
decoder fuses current evidence, temporally aggregated intention context and
predicted dynamics,
\begin{equation}
\mathbf{u}_t=[\mathbf{S}_t\;\|\;\mathbf{c}_t\;\|\;\mathbf{w}_t],\qquad
\boldsymbol{\ell}_t=C_{\omega}(\mathbf{u}_t)\in\mathbb{R}^{NM},
\end{equation}
and predicts
$\hat{h}_t=\arg\max_{i}\boldsymbol{\ell}_{t,i}$.

Because the labels live on a 2-D grid, a coordinate consistency term exploits
the spatial structure. For class index $i\in\{0,\ldots,NM-1\}$, we define its
row and column coordinates as
$\rho_i=\lfloor i/M\rfloor$ and $\kappa_i=i\bmod M$.
For the ground-truth class $h_t$, let
$\rho_t^{\star}=\rho_{h_t}$ and $\kappa_t^{\star}=\kappa_{h_t}$.
With $\mathbf{p}_t=\mathrm{softmax}(\boldsymbol{\ell}_t)$, the expected
coordinates are
$\hat{\rho}_t=\sum_i \mathbf{p}_{t,i}\rho_i$ and
$\hat{\kappa}_t=\sum_i \mathbf{p}_{t,i}\kappa_i$,
\begin{equation}
\mathcal{L}_{\mathrm{coord}}=
\mathrm{SmoothL1}(\hat{\rho}_t,\rho_t^{\star})
+\mathrm{SmoothL1}(\hat{\kappa}_t,\kappa_t^{\star}).
\end{equation}

We retain a weak predictive-consistency term during fine-tuning to encourage preservation of the learned transition structure. We define the retention loss
using the same predictive weights as in Stage~1,
\begin{equation}
\mathcal{L}_{\mathrm{keep}}
=
\lambda_{\mathrm{latent}}\mathcal{L}_{\mathrm{latent}}
+
\lambda_{\mathrm{cos}}\mathcal{L}_{\mathrm{cos}},
\end{equation}
where $\lambda_{\mathrm{latent}}=1.0$ and
$\lambda_{\mathrm{cos}}=0.1$ are shared with Stage~1.

The complete Stage~2 objective is
\begin{equation}
\mathcal{L}_{\mathrm{stage2}}=
\lambda_{\mathrm{cls}}\mathcal{L}_{\mathrm{cls}}
+\lambda_{\mathrm{coord}}\mathcal{L}_{\mathrm{coord}}
+\lambda_{\mathrm{keep}}\mathcal{L}_{\mathrm{keep}},
\end{equation}
where $\mathcal{L}_{\mathrm{cls}}=
\mathrm{CE}(\boldsymbol{\ell}_t,h_t)$. Unless otherwise stated, we use
$\lambda_{\mathrm{cls}}=1.0$,
$\lambda_{\mathrm{coord}}=0.1$, and
$\lambda_{\mathrm{keep}}=0.03$.
The future EEG window is used only to compute
$\mathcal{L}_{\mathrm{keep}}$ during training and is never required at
inference. On datasets without a 2-D label geometry
(Section~\ref{sec:setup}), $\lambda_{\mathrm{coord}}$ is set to zero and
$NM$ is replaced by the number of classes.

\subsection{Robot Target Selection}
\label{sec:robot_selection}
The user selects one of four candidate objects without speech, gesture, or manual command. No dedicated eye tracker is used. For each robot-camera observation,
SAM \cite{kirillov2023segment} first segments the four objects present in the
scene,
$\mathcal{O}=\{o_1,\ldots,o_J\}$ with $J=4$. The image plane is divided into
the same $6\times8$ grid used by EEG-VID, and each segmented object is mapped
to a grid label $g(o_j)$. SAM therefore defines the scene-dependent candidate
label set
\begin{equation}
\mathcal{C}=\{g(o_1),\ldots,g(o_J)\}, \qquad J=4.
\end{equation}
EEG-VID independently produces logits
$\boldsymbol{\ell}_{t}\in\mathbb{R}^{48}$ from EEG. The final target is the
object whose SAM-derived grid label receives the largest EEG logit,
\begin{equation}
\hat{o}=\arg\max_{o_j\in\mathcal{O}}\boldsymbol{\ell}_{t,g(o_j)}.
\label{eq:sam_select}
\end{equation}
SAM supplies only the four candidate spatial labels. It does not infer the user's intended target. EEG-VID makes the intention decision by restricting its posterior to those labels. Because every evaluated scene contains four candidate objects, random selection within the candidate set has a 25\% chance level.

We evaluate this decision rule in an offline robot-scene experiment using six
tabletop configurations, each containing four objects at different spatial
positions. For each RGB observation, SAM produces the object masks, and each
object is assigned to the grid cell containing the largest number of pixels
from its mask. The subject-specific EEG-VID model supplies the 48-way spatial
scores, after which \eqref{eq:sam_select} selects among the four occupied
candidate cells.

The selected mask can be fused with depth for downstream grasp planning
(AnyGrasp \cite{fang2023anygrasp}). The reported outcome is
\emph{target selection} before motion planning. Grasp execution is not evaluated.

\subsection{Data Acquisition, Ethics, and Preprocessing}
\label{sec:ethics}

\subsubsection{Ethics statement}
Seven healthy adults participated in total. One contributed the VIG-48 cross-day corpus, and six contributed the robot-scene set. All participants
provided written informed consent.

\subsubsection{Acquisition and preprocessing}

\begin{figure}[!t]
\centering
\includegraphics[width=0.45\columnwidth]{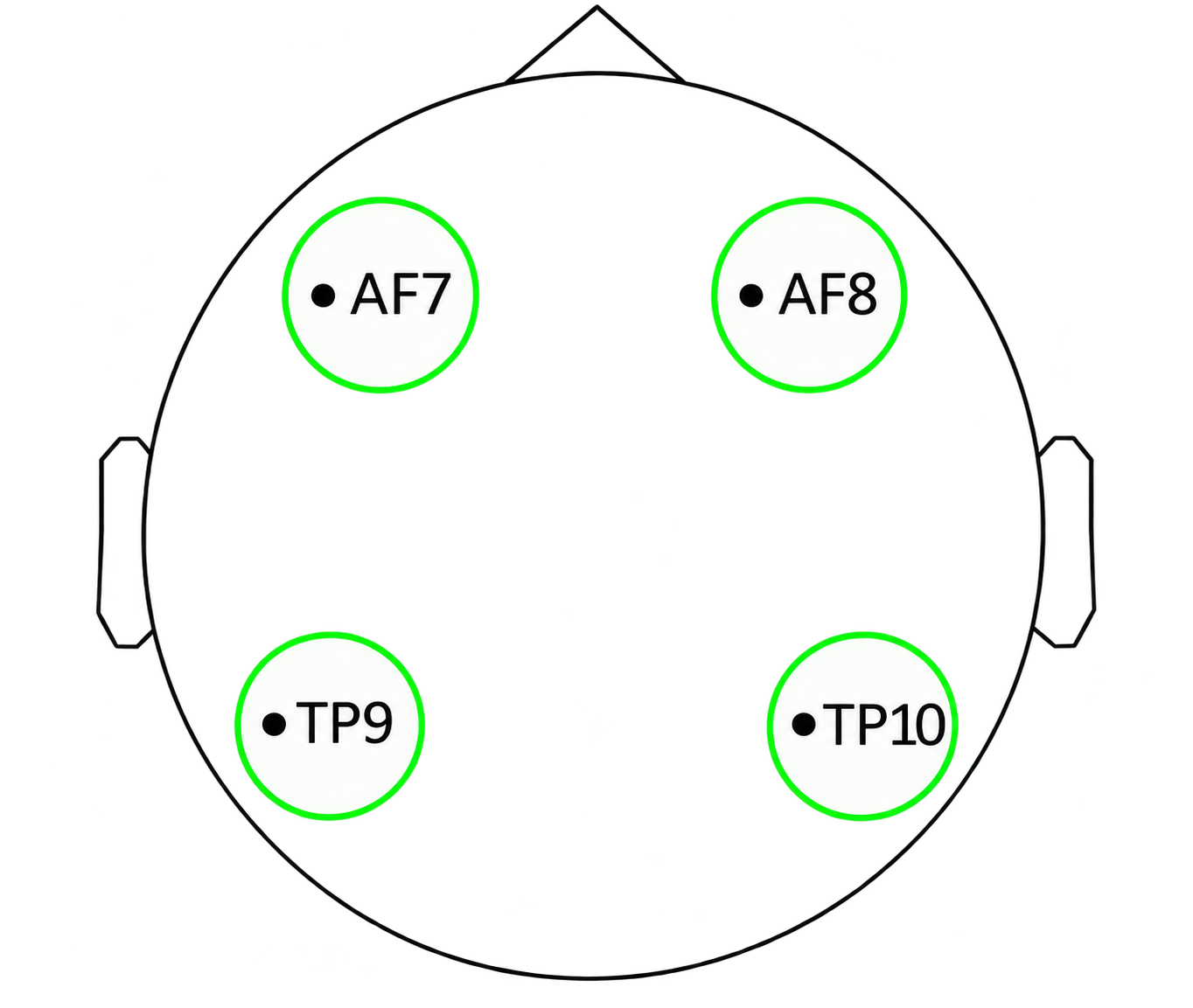}
\caption{Four-electrode Muse S Athena montage used for the in-house recordings at AF7, AF8, TP9, and TP10. The layout is shown for acquisition
reproducibility and is not used for source-localization claims.}
\label{fig:muse_electrodes}
\end{figure}

EEG was recorded at 256\,Hz with a Muse S Athena using AF7, AF8, TP9, and TP10.
Participants viewed a $6\times8$ grid. A target cell at a randomly selected position was continuously highlighted for 3--6\,s. The target did not flicker, so no
frequency-tagged SSVEP cue was used. No dedicated eye tracker and no
gaze-contingent hardware were used at any stage. Participants were not instructed to fixate centrally. Under this single-device protocol, natural orienting toward the intended cell remains part of the recorded input rather than being explicitly suppressed. The consequences of this choice for
what can be claimed about the decoded signal are stated in
Section~\ref{sec:origin}.

Recordings were segmented into non-overlapping 1-s windows. A fixed
preprocessing pipeline---non-finite-value handling, outlier suppression,
fourth-order 0.5--45\,Hz band-pass filtering, and temporal smoothing---was
applied consistently across all in-house experiments. No ICA or dedicated ocular-artifact rejection was applied. With four scalp channels and no reference EOG, ocular components cannot be identified reliably. Removing orienting-related activity could also suppress information available to the device-level interface. We
therefore retain a fixed preprocessing pipeline and interpret VIG-48 at the
device level rather than as cortical source decoding.

\section{Experimental Setup}
\label{sec:setup}

\subsection{Datasets}

To evaluate decoding performance and transferability, we use the in-house
VIG-48 dataset and two public BCI Competition IV benchmarks. We further
conduct a separate robot-scene experiment to assess scene-informed target
selection.

\subsubsection{In-house EEG visual intention dataset (VIG-48)}

VIG-48 contains approximately 15{,}000 window-level samples from one participant recorded across multiple days. Approximately 13{,}000 samples are used for model development, and the held-out test split is collected on three days that do not overlap the training days, so that evaluation measures cross-day distribution shift rather than within-day generalization. Within the model-development portion, training and validation are separated by complete event identifiers before Stage~1 context construction. All windows from a given cued event remain in one split. The task is region-level decoding over a $6\times8$ grid with 48 candidate regions. The Top-1 and Top-5 chance levels are $2.08\%$ and $10.42\%$, respectively. Because
multiple windows originate from the same cued trial, individual windows are
not treated as independent observations for a binomial significance test.

\subsubsection{Robot-scene target-selection set}
A separate in-house robot-scene set is collected from six additional
participants across six tabletop scene configurations. Every test scene
contains four candidate objects whose spatial positions vary across
configurations. Before robot-scene evaluation, each participant completes 48
subject-specific calibration trials using the standard visual-intention
acquisition paradigm. Each calibration trial lasts 10\,s with one target
label, providing 480\,s of calibration EEG per participant. Robot-scene test
trials are collected separately and last 30\,s. During testing, each
participant views the scene from the robot-camera viewpoint. RGB and depth
images from this viewpoint are recorded together with the EEG trials. Seven
robot-scene test trials are excluded because of data-saving or processing
failures.

For each participant, EEG-VID is initialized from the visual-intention model
and adapted using only that participant's calibration data. This adaptation
is performed in Stage~2 with supervised classification fine-tuning; Stage~1
is not re-run on the robot calibration set. The adapted model is then
evaluated on the participant's held-out robot-scene trials. The reported
endpoint is window-level candidate-constrained target-selection accuracy,
defined as the number of windows for which the correct object's grid label
has the largest EEG score among the four SAM-derived candidate labels divided
by the total number of valid test windows.

\subsubsection{Public benchmarks}
To test whether the proposed pretraining principle transfers beyond our
acquisition setup, we evaluate on BCI Competition IV dataset~2a
\cite{brunner2008bci2a} and dataset~2b \cite{leeb2008bci2b}, released as part
of BCI Competition IV \cite{tangermann2012review}. IV-2a is a
four-class motor-imagery benchmark with 22 EEG channels, and IV-2b is a
two-class motor-imagery benchmark with three EEG channels. EOG channels are
excluded in both datasets. These datasets are not visual-intention tasks. They are included to test whether Stage~1 transfers to standard EEG decoding settings. Chance levels are 25\% for IV-2a and 50\% for IV-2b. The three evaluation protocols applied to them are defined in
Section~\ref{sec:protocols}.

\subsection{Evaluation Protocols}
\label{sec:protocols}
Absolute accuracy on IV-2a and IV-2b depends strongly on how models are
trained and how decisions are made. Rather than reporting a single
configuration, we evaluate every method under three protocols that differ in
the amount of subject-specific data available at training time. All three use
the same preprocessing, 1-s windowing, optimizer settings, and the same
five fixed random seeds. In all three protocols, the classification
decision is made per 1-s window rather than per trial.

For every protocol, raw trials/events retain an event identifier through
preprocessing and windowing. Training and validation are then separated by
complete event identifiers, so all windows from one event remain in the same
split. Data-dependent normalization statistics are estimated from training
windows only and applied unchanged to validation and test data. Stage~1
history--target contexts are constructed only after this split and
independently within each split and each event; no context crosses an event
boundary or a train/validation/test boundary.

\subsubsection{Pooled cross-session (subject-overlapping)}
A single subject-agnostic model is trained on the pooled training sessions of
all nine subjects and evaluated on the pooled independent evaluation
sessions of the same nine subjects. For IV-2a, the training pool is A01T--A09T and the test set is A01E--A09E. For IV-2b, the training pool contains Bxx01T, Bxx02T, and Bxx03T, while the test set contains Bxx04E and Bxx05E (xx$=01,\ldots,09$). Training and testing therefore share subjects but not
sessions. For each seed, 20\% of the pooled training events are held out for
validation ($\mathrm{val\_ratio}=0.2$), with all windows from an event kept together. Evaluation sessions are never used for model selection. After segmentation, IV-2a contains 10,368 training-pool
windows and 10,368 test windows, and IV-2b contains 14,720 and 11,360. Each
IV-2a trial contributes exactly four consecutive 1-s windows
($9\ \text{subjects}\times288\ \text{trials}\times4=10{,}368$), taken from
the fixed 4-s motor-imagery interval. This protocol measures session shift
with subject identity held fixed and is used for the main comparison and for
all ablations.

\subsubsection{Within-subject}
One model is trained per subject on that subject's own training session(s)
and evaluated on that subject's own evaluation session(s), with no data from
any other subject. The validation subset is drawn at the event level from
that subject's training session(s), never from the evaluation session(s).
This matches the conventional way IV-2a and IV-2b are reported, except that
decisions remain window-level. It measures how each method behaves when
subject-specific data are available but scarce.

\subsubsection{Cross-subject (leave-one-subject-out)}
For each held-out subject, the model is trained on the remaining eight
subjects and evaluated on the held-out subject's sessions, with no
calibration data from that subject. Validation events are selected only from the eight training subjects. The held-out subject is never used for normalization, model selection, or checkpoint selection. This is the strictest of the three and the one most relevant to a deployable interface,
since a new user supplies no labelled data.

\subsubsection{Interpreting the absolute values}
\label{sec:protocol_note}
All IV-2a/IV-2b results are 1-s-window accuracies, whereas standard
competition protocols classify complete trials. We also do not use
sliding-window augmentation, filter-bank features, or covariance alignment
\cite{He2018TransferLF}. Absolute values therefore should not be compared
directly with published competition results. Because all methods share the
same preprocessing and decision protocol, the within-table comparisons test
the training strategy under a common controlled setting.

\subsection{Window Construction and Temporal Context}
\label{sec:window_protocol}
All datasets are segmented into consecutive, non-overlapping 1-s EEG windows.
IV-2a/IV-2b contain 250 samples per window (250~Hz), whereas the in-house data
contain 256 samples (256\,Hz). A fourth-order 0.5--45\,Hz band-pass filter is
used throughout. We fix the history length to $K=4$.

For consecutive windows $\{\mathbf{x}_0,\mathbf{x}_1,\ldots\}$, Stage~1 uses
\begin{equation}
[\mathbf{x}_{t-3},\mathbf{x}_{t-2},\mathbf{x}_{t-1},\mathbf{x}_{t}]
\longrightarrow \mathbf{x}_{t+1}.
\end{equation}
Thus, each prediction uses at most 4\,s of history to predict the next
1-s window. Missing history at the start of an event is left-padded by
repeating the first window. Raw windows do not overlap, although adjacent
history--target samples from the same event share three historical windows.
This overlap occurs only within a single event and a single data split. Stage~1 contexts are generated after event-level splitting and never cross event or train/validation/test boundaries. The terminal window of each event
is omitted whenever a future target is required.

For IV-2a and IV-2b, where an event contributes only four windows, this
scheme yields three Stage~1 samples per trial and the first two are partly left-padded. The effective history is therefore shorter than 4\,s on these benchmarks. On the in-house recordings, where each highlighted target
lasts 3--6\,s, full-length histories are available for most samples. We report
this asymmetry because it may explain part of the variation in Stage~1 gains
across datasets.

\subsection{Baselines, Variants and Metrics}
We compare EEG-VID with EEGNet \cite{lawhern2018eegnet}, DeepConvNet
\cite{Schirrmeister_2017}, EEG-Conformer \cite{song2022eeg}, TSConv
\cite{song2024decoding}, Neuro-3D \cite{guo2025neuro}, EEG2Rep
\cite{foumani2024eeg2rep}, and BENDR \cite{kostas2021bendr}. EEG-VID w/o Stage~1 retains the supervised
EEG-VID architecture but omits predictive pretraining. All reported baseline
numbers are produced by our local implementations under the common data
splits and preprocessing pipeline rather than copied from the cited papers.
Unless stated otherwise, results use five fixed random seeds and are reported as mean~\mypm~standard deviation.
We report Top-1 accuracy on all datasets and Top-5 accuracy on the 48-way
in-house task. The public-benchmark accuracies are reported at the 1-s-window
level under each of the three protocols of Section~\ref{sec:protocols}. For the
robot-scene experiment we report SAM-constrained target-selection accuracy,
i.e., the fraction of held-out test cases for which the highest-logit
candidate object matches the intended object. With four candidates, random
chance is 25\%.

\subsection{Stage~1 Transfer to Existing EEG Backbones}
\label{sec:backbone_transfer}
To separate the training objective from the EEG-VID encoder, Stage~1 is also
applied to EEGNet, DeepConvNet, EEG-Conformer, Neuro-3D and TSConv. For each
backbone $B$, the classifier output is replaced during Stage~1 by a latent
adaptation layer, giving $\mathbf{z}_t=B(\mathbf{x}_t)$. An EMA copy of the
same backbone provides the stopped-gradient future target
$\mathbf{z}^{\mathrm{target}}_{t+1}$, and a predictor estimates that target
from the $K$ historical latent states. A lightweight auxiliary classifier
uses the historical representation and the same training labels as EEG-VID.
Thus, the external-backbone variants use the same three Stage~1 loss components and coefficients in \eqref{eq:stage1}, including latent distance, cosine consistency, and weak task guidance.

After Stage~1, the pretrained backbone initializes its original supervised
decoder and is fine-tuned under the same Stage~2 data split and optimizer
policy as the corresponding baseline. We denote the variants pretrained with task-guided latent prediction by the prefix TLP, giving TLP-EEGNet, TLP-DeepConvNet, TLP-Conformer, TLP-Neuro3D, and TLP-TSConv. These variants are not treated as additional competitors in the EEG-VID architecture ranking. Instead, each TLP variant is compared with its corresponding backbone without Stage~1 to test whether the training strategy transfers.

To further separate the Stage~1 training principle from the proposed
predictor architecture, we replace the predictor with LSTM, GRU, and standard Transformer alternatives for EEG-VID and all five TLP variants under the pooled protocol. All replacements use the same latent input/output dimensions,
Stage~1 objective, task-guidance weight, data splits, and training policy, with
no predictor-specific hyperparameter tuning.

\subsection{Statistical Analysis}
\label{sec:stats}
Results from repeated optimization runs are reported as
mean~\mypm~standard deviation over five matched seeds. Seeds quantify
optimization variability and are not treated as independent participants.
For the within-subject and leave-one-subject-out protocols, the primary
summary is the mean across the nine subject-specific evaluations.
Table~\ref{tab:subject_summary} additionally reports the number of subjects
whose mean accuracy improves after Stage~1 ($k/9$), exposing the paired
direction of the effect without treating those counts as a population-level
significance test.

We also report the direction of the accuracy change across matched comparisons. Because these comparisons reuse datasets, subjects, seeds, and
related model families, the resulting counts are not treated as independent
observations or population-level significance tests.
Table~\ref{tab:predictor_ablation} additionally reports the mean and standard
deviation of paired accuracy differences across matched model--dataset
settings. For pooled backbone comparisons using matched seeds,
Table~\ref{tab:wm_backbone} reports the paired change directly.

\subsection{Implementation Details}
All models are trained with AdamW ~\cite{loshchilov2019decoupled} on a single NVIDIA RTX~4090 using a batch
size of 64. Training uses validation-based early stopping with a patience of
100 and a maximum allowance of 10{,}000 epochs. For the training-budget
sensitivity control in Section~\ref{sec:transfer}, all six corresponding single-stage models are additionally rerun with a fourfold larger maximum allowance
under the same early-stopping rule. Weight decay is $10^{-4}$ for both stages;
learning rates are $3\times10^{-4}$ for Stage~1 and $1\times10^{-4}$ for
Stage~2. Unless changed by an ablation, Stage~1 uses
$\lambda_{\mathrm{latent}}=1.0$, $\lambda_{\mathrm{cos}}=0.1$,
$\lambda_{\mathrm{cls}}^{S1}=0.1$, and EMA decay 0.995. Stage~2 uses
$\lambda_{\mathrm{cls}}=1.0$, $\lambda_{\mathrm{coord}}=0.1$, and
$\lambda_{\mathrm{keep}}=0.03$, with cosine coefficient 0.1 inside
$\mathcal{L}_{\mathrm{keep}}$. The robot target-selection formulation uses
candidate objects segmented from RGB-D observations and matches them to the
decoded spatial region as described in the Methods section.


\section{Results}

\subsection{Pooled Cross-Session Comparison}

\begin{table*}[!t]
\caption{Pooled cross-session comparison. Best results are shown in
\textbf{bold}.}
\label{tab:main_results}
\centering
\begin{threeparttable}
\footnotesize
\renewcommand{\arraystretch}{1.08}
\begin{tabular*}{0.82\textwidth}{@{\extracolsep{\fill}}lcccc@{}}
\toprule
\multirow{2}{*}{Method}
& \multicolumn{2}{c}{VIG-48 (48-way)$^\dagger$}
& \multirow{2}{*}{IV-2a$^{\dagger\ddagger}$}
& \multirow{2}{*}{IV-2b$^{\dagger\ddagger}$}\\
\cmidrule(lr){2-3}
& Top-1 & Top-5 & & \\
\midrule
EEGNet~\cite{lawhern2018eegnet} & 4.80\mypm0.11 & 22.74\mypm1.02 & 45.45\mypm1.08 & 70.01\mypm0.19\\
DeepConvNet~\cite{Schirrmeister_2017} & 4.35\mypm0.45 & 21.46\mypm1.16 & 50.02\mypm0.18 & 70.48\mypm0.40\\
EEG-Conformer~\cite{song2022eeg} & 4.95\mypm0.18 & 21.98\mypm0.24 & 49.15\mypm0.51 & 69.07\mypm1.00\\
Neuro-3D~\cite{guo2025neuro} & 4.57\mypm0.31 & 19.95\mypm1.47 & 41.87\mypm0.91 & 68.10\mypm0.53\\
TSConv~\cite{song2024decoding} & 3.67\mypm0.30 & 16.48\mypm0.83 & 29.24\mypm0.50 & 52.27\mypm0.50\\
\midrule
EEG2Rep~\cite{foumani2024eeg2rep} & 3.43\mypm0.11 & 16.54\mypm1.18 & 32.62\mypm1.02 & 67.17\mypm1.02\\
BENDR~\cite{kostas2021bendr} & 3.25\mypm0.32 & 13.92\mypm0.68 & 30.24\mypm1.14 & 66.47\mypm0.72\\
\midrule
EEG-VID w/o Stage~1 & 5.05\mypm0.72 & 20.67\mypm3.27 & 40.00\mypm1.10 & 67.67\mypm0.91\\
EEG-VID & \textbf{6.52}\mypm0.95 & \textbf{30.50}\mypm1.79
& \textbf{51.30}\mypm0.42 & \textbf{75.90}\mypm1.23\\
\bottomrule
\end{tabular*}
\begin{tablenotes}[flushleft]
\scriptsize
\item[] $\dagger$: Values are mean~\mypm~standard deviation over five fixed
random seeds under the shared local pipeline.
\item[] $\ddagger$: IV-2a and IV-2b report 1-s-window accuracy.
\end{tablenotes}
\end{threeparttable}
\end{table*}

Table~\ref{tab:main_results} compares EEG-VID with seven locally retrained
methods and its no-Stage~1 ablation. 
Under the pooled protocol, EEG-VID achieves $6.52\pm0.95\%$/ $30.50\pm1.79\%$ Top-1/Top-5 on VIG-48, $51.30\pm0.42\%$ on IV-2a, and $75.90\pm1.23\%$ on IV-2b. Relative to the identical architecture without Stage~1, Top-1 improves by 1.47, 11.30, and
8.23 points, respectively, isolating the effect of the two-stage training
procedure from decoder architecture.

EEG2Rep and BENDR provide direct predictive/self-supervised reference points.
Under the same local evaluation pipeline, both are below EEG-VID on the three
Top-1 tasks and on VIG-48 Top-5. Because both were designed for larger
montages and larger pretraining corpora, this comparison should be read as
evidence about behaviour in the present low-channel, small-data regime rather
than as a general ranking of the methods.

\subsection{Stage~1 Transfer across EEG Backbones}
\label{sec:transfer}

\begin{table}[!t]
\caption{Stage~1 gains across EEG backbones.}
\label{tab:wm_backbone}
\centering
\begin{threeparttable}
\footnotesize
\setlength{\tabcolsep}{2.8pt}
\renewcommand{\arraystretch}{1.10}
\begin{tabularx}{\columnwidth}{@{}>{\raggedright\arraybackslash}Xcccc@{}}
\toprule
& \multicolumn{2}{c}{VIG-48$^\dagger$}
& IV-2a$^\dagger$ & IV-2b$^\dagger$\\
\cmidrule(lr){2-3}
Model pair & Top-1 & Top-5 & Top-1 & Top-1\\
\midrule
TLP-EEGNet / EEGNet & +1.60\mypm0.68 & +4.96\mypm2.53 & +11.20\mypm1.29 & +6.41\mypm1.04\\
TLP-DeepConvNet / DeepConvNet & +1.28\mypm1.26 & +3.50\mypm2.11
& +19.29\mypm1.00 & +7.09\mypm1.00\\
TLP-Conformer / Conformer & +1.24\mypm1.27 & +4.06\mypm2.28
& +6.38\mypm0.87 & +6.92\mypm1.33\\
TLP-Neuro3D / Neuro-3D & +2.93\mypm1.65 & +7.63\mypm1.57
& +7.85\mypm1.20 & +6.86\mypm0.87\\
TLP-TSConv / TSConv & +0.65\mypm0.42 & +0.82\mypm3.69
& +1.95\mypm0.80 & +0.05\mypm1.16\\
\midrule
EEG-VID / w/o Stage~1$^\ddagger$
& +1.47\mypm1.19
& +9.83\mypm3.73
& +11.30\mypm1.18
& +8.23\mypm1.53\\
\bottomrule
\end{tabularx}
\begin{tablenotes}[flushleft]
\scriptsize
\item[] $\dagger$: Entries are seed-paired accuracy changes in percentage
points after Stage~1 relative to the matched backbone, reported as
mean~\mypm~standard deviation under the pooled protocol.
\item[] $\ddagger$: EEG-VID is compared with the identical architecture
without Stage~1.
\end{tablenotes}
\end{threeparttable}
\end{table}

Table~\ref{tab:wm_backbone} addresses a different question by comparing each decoder only with itself before and after Stage~1. All 15 external-backbone Top-1 mean changes are non-negative, and all five backbones also improve on VIG-48 Top-5. The magnitude of the gain depends on the architecture. For example, DeepConvNet gains $19.29$ points on IV-2a whereas TSConv changes by only $0.05\pm1.16$ points on IV-2b. The latter is best interpreted as no measurable
effect.

Some Stage~1-enhanced external backbones exceed EEG-VID in absolute accuracy
(e.g., TLP-Neuro3D on VIG-48 and TLP-DeepConvNet on IV-2a/IV-2b). This result
further supports interpreting Stage~1 as a transferable training strategy
rather than an architecture-specific advantage. EEG-VID is an application-oriented model with a grid-structured decoder that maps decoded regions to candidate objects. The gain also varies across architectures, showing that Stage~1 does not provide a fixed improvement across all models and datasets.

As a training-budget sensitivity control, all six single-stage counterparts
were rerun with a fourfold larger maximum epoch allowance under the same
early-stopping rule. Stage~1 remained higher in 23 of 24 pooled model--metric comparisons. The only exception was TLP-TSConv on IV-2b, where the difference was $-0.07$ percentage points. Thus, the maximum epoch limit of the single-stage models does not explain the main pooled gains, although this control does not exactly match the number of parameter
updates or total computation.

\subsection{Generalization across Subject-Wise Protocols}
\label{sec:subject_protocols}

\begin{table*}[!t]
\caption{Subject-wise public-benchmark comparison. Best results are shown in
\textbf{bold}.}
\label{tab:subject_summary}
\centering
\begin{threeparttable}
\footnotesize
\renewcommand{\arraystretch}{1.10}
\begin{tabular*}{0.94\textwidth}{@{\extracolsep{\fill}}lcccc@{}}
\toprule
Model & IV-2a Within$^\dagger$ & IV-2b Within$^\dagger$
& IV-2a LOSO$^\dagger$ & IV-2b LOSO$^\dagger$\\
\midrule
TLP-EEGNet & 53.38 (+0.40; 6/9) & 75.57 (+3.88; 6/9)
& 45.24 (+8.30; 9/9) & 70.54 (+3.02; 7/9)\\
TLP-DeepConvNet & \textbf{56.71} (+5.59; 7/9)
& \textbf{76.52} (+6.36; 8/9)
& \textbf{54.19} (+16.22; 9/9)
& \textbf{73.08} (+5.69; 7/9)\\
TLP-Conformer & 55.11 (+6.59; 8/9) & 74.58 (+5.27; 8/9)
& 42.04 (+8.29; 8/9) & 70.45 (+3.26; 6/9)\\
TLP-Neuro3D & 49.49 (+2.86; 6/9) & 73.20 (+5.32; 6/9)
& 41.82 (+7.59; 8/9) & 69.44 (+3.68; 7/9)\\
TLP-TSConv & 32.42 (+1.68; 5/9) & 52.73 ($-0.24$; 5/9)
& 28.37 (+0.93; 7/9) & 51.72 (+1.70; 8/9)\\
EEG-VID & 48.76 (+6.08; 8/9) & 73.18 (+6.31; 8/9)
& 40.80 (+7.25; 9/9) & 70.69 (+5.50; 8/9)\\
\bottomrule
\end{tabular*}
\begin{tablenotes}[flushleft]
\scriptsize
\item[] $\dagger$: Entries are accuracy (\%) followed by
($\Delta$; $k/9$), where $\Delta$ denotes the mean change versus the matched
non-Stage~1 backbone and $k$ the number of subjects with a positive paired
change. ``Within'' uses subject-specific training; ``LOSO'' uses no labelled
data from the evaluated subject.
\end{tablenotes}
\end{threeparttable}
\end{table*}

The subject-wise protocols separate absolute model ranking from Stage~1
transfer. Under within-subject evaluation, Stage~1 yields positive mean changes for 11 of 12 model--dataset pairs. The only negative mean is TLP-TSConv on IV-2b at $-0.24$ points. The $k/9$ counts in Table~\ref{tab:subject_summary} show that the improvements are observed across multiple participants. EEG-VID improves for 8/9 subjects on
both IV-2a and IV-2b, while TLP-DeepConvNet and TLP-Conformer improve for at
least 7/9 and 8/9 subjects, respectively.

The strongest pattern appears under leave-one-subject-out (LOSO) evaluation.
Stage~1 improves the mean for all six models on both public datasets. On IV-2a, EEG-VID, TLP-EEGNet, and TLP-DeepConvNet improve for all 9/9 held-out subjects. TLP-Conformer and TLP-Neuro3D improve for 8/9. The largest mean gain is
+16.22 points for DeepConvNet, while EEG-VID improves from 33.55\% to 40.80\%.
On IV-2b, all six models again have positive mean changes, with subject-level
improvement counts ranging from 6/9 to 8/9. The LOSO results show that the Stage~1 benefit extends beyond subject-overlapping cross-session evaluation.

Across pooled, within-subject, and LOSO evaluations, Stage~1 produces a
positive mean change in 41 of 42 matched decoder--dataset--protocol
comparisons. This count and the $k/9$ entries are descriptive rather than
population-level significance tests (Section~\ref{sec:stats}).

\subsection{Stage~1 Objective Ablation}
\label{sec:stage1_ablation}

\begin{table}[!t]
\caption{Effect of weak task guidance.}
\label{tab:stage1_ablation}
\centering
\begin{threeparttable}
\footnotesize
\setlength{\tabcolsep}{3.5pt}
\renewcommand{\arraystretch}{1.10}
\begin{tabularx}{\columnwidth}{@{}>{\raggedright\arraybackslash}Xcccc@{}}
\toprule
Model
& VIG-48$^\dagger$
& IV-2a$^\dagger$
& IV-2b$^\dagger$
& 0.1 Best/3$^\ddagger$\\
\midrule
EEG-VID & +0.62 & +2.83 & +2.55 & 3\\
TLP-EEGNet & +1.19 & +3.57 & +1.95 & 3\\
TLP-DeepConvNet & +0.02 & +1.18 & +1.32 & 2\\
TLP-Conformer & +0.42 & +1.61 & +1.71 & 3\\
TLP-Neuro3D & +1.44 & +2.26 & +0.03 & 3\\
TLP-TSConv & +1.19 & +0.61 & +0.59 & 2\\
\bottomrule
\end{tabularx}
\begin{tablenotes}[flushleft]
\scriptsize
\item[] $\dagger$: Entries are
$\mathrm{Acc}_{0.1}-\mathrm{Acc}_{0}$ in percentage points under the pooled
protocol.
\item[] $\ddagger$: Number of datasets for which 0.1 gives the best of the three nonzero weights $\{0.1,0.2,0.3\}$.
\end{tablenotes}
\end{threeparttable}
\end{table}

The default weak task-guidance setting ($\lambda_{\mathrm{cls}}^{S1}=0.1$) exceeds latent-only pretraining in all 18 model--dataset comparisons
(Table~\ref{tab:stage1_ablation}). For EEG-VID, latent-only pretraining
obtains 5.90\%, 48.47\% and 73.35\% on VIG-48, IV-2a and IV-2b, compared
with 6.52\%, 51.30\% and 75.90\% for the default. Latent-only EEG-VID also outperforms the no-Stage~1 model on all three datasets, indicating that latent prediction alone provides consistent gains even without weak task guidance.

Weights 0.2 and 0.3 underperform the default in 34 of 36 comparisons,
except for TLP-DeepConvNet on VIG-48 (0.3) and TLP-TSConv on IV-2a (0.2). Thus, a small nonzero task-guidance weight is the most reliable choice within the tested range. These results do not show that it is globally optimal.

\subsection{Predictor Architecture Robustness}
\begin{table}[!t]
\caption{Predictor-architecture robustness.}
\label{tab:predictor_ablation}
\centering
\begin{threeparttable}
\scriptsize
\renewcommand{\arraystretch}{1.10}
\begin{tabular*}{\columnwidth}{@{\extracolsep{\fill}}lccc@{}}
\toprule
Replacement
& S1 $>$ DS$^\dagger$
& Proposed $>$ repl.$^\dagger$
& $\Delta$Acc.$^\ddagger$\\
\midrule
LSTM
& 16/18
& 16/18
& +0.64\mypm0.67\\
GRU
& 16/18
& 13/18
& +0.52\mypm0.70\\
Transformer
& 16/18
& 16/18
& +0.52\mypm0.52\\
\midrule
Overall
& \textbf{48/54}
& \textbf{45/54}
& \textbf{+0.56}\mypm\textbf{0.63}\\
\bottomrule
\end{tabular*}
\begin{tablenotes}[flushleft]
\scriptsize
\item[] $\dagger$: Counts use 18 matched model--dataset settings per predictor
replacement under the pooled protocol (54 overall); S1 denotes Stage~1 and
DS direct supervision.
\item[] $\ddagger$: $\Delta\mathrm{Acc}
=\mathrm{Acc}_{\mathrm{proposed}}
-\mathrm{Acc}_{\mathrm{replacement}}$, reported as
mean~\mypm~standard deviation in percentage points.
\end{tablenotes}
\end{threeparttable}
\end{table}

Table~\ref{tab:predictor_ablation} tests whether the Stage~1 benefit depends
on the proposed predictor. LSTM, GRU, and Transformer replacements retain an
advantage over direct supervision in 48 of 54 pooled model--dataset
comparisons, including 18 of 18 on IV-2a. Thus, the Stage~1 benefit is not
tied to a specific predictor architecture. The proposed predictor is
nevertheless higher in 45 of 54 matched comparisons, with an overall mean
advantage of 0.56\mypm0.63 percentage points. These results separate the
transferable Stage~1 training principle from the additional,
architecture-dependent benefit of the proposed predictor.

\subsection{Electrode-Subset Analysis}

\begin{table}[!t]
\caption{Electrode-subset analysis on VIG-48.}
\label{tab:channels}
\centering
\begin{threeparttable}
\footnotesize
\setlength{\tabcolsep}{5pt}
\renewcommand{\arraystretch}{1.08}
\begin{tabular}{@{}lccc@{}}
\toprule
\multirow{2}{*}{Method}
& \multicolumn{3}{c}{Top-1 accuracy (\%)$^\dagger$}\\
\cmidrule(lr){2-4}
& Four electrodes & AF7/AF8 & TP9/TP10\\
\midrule
EEG-VID & 6.52\mypm0.95 & 5.84\mypm1.11 & 4.70\mypm0.92\\
EEG-VID w/o Stage~1 & 5.05\mypm0.72 & 4.27\mypm0.67 & 3.59\mypm0.19\\
EEGNet & 4.80\mypm0.11 & 4.20\mypm0.34 & 3.40\mypm0.29\\
DeepConvNet & 4.35\mypm0.45 & 4.16\mypm0.22 & 3.16\mypm0.37\\
EEG-Conformer & 4.95\mypm0.18 & 4.19\mypm0.64 & 3.59\mypm0.26\\
Neuro-3D & 4.57\mypm0.31 & 4.38\mypm0.23 & 3.26\mypm0.36\\
TSConv & 3.67\mypm0.30 & 3.60\mypm0.29 & 3.09\mypm0.30\\
\bottomrule
\end{tabular}
\begin{tablenotes}[flushleft]
\scriptsize
\item[] $\dagger$: Values are mean~\mypm~standard deviation over five fixed
random seeds. Comparisons are descriptive within each model and do not imply
source localization.
\end{tablenotes}
\end{threeparttable}
\end{table}

For EEG-VID, the full montage is best, followed by AF7/AF8 and then
TP9/TP10. The same frontal-over-temporo-parietal ordering appears for the
five supervised baselines. Because AF7/AF8 are close to the eyes and gaze was
not constrained, this pattern cannot be interpreted as cortical
localization. It only shows that the frontal pair carries stronger
standalone target-related information in the present device configuration.

\subsection{Offline Robot-Scene Target Selection}
\label{sec:robot_results}

We next evaluate whether the decoded 48-region posterior can support
object-level selection when scene information restricts the decision space.
For every robot-camera frame, SAM identifies the four object masks and maps
them to four candidate grid labels. The subject-specific adapted EEG-VID
model is applied to the held-out robot-scene EEG, and its 48-way scores are
restricted to these four SAM-derived labels using
\eqref{eq:sam_select}. SAM therefore determines which spatial labels are
eligible candidates, whereas EEG-VID determines which candidate is selected.

\begin{figure}[!t]
    \centering
    \includegraphics[width=\columnwidth]{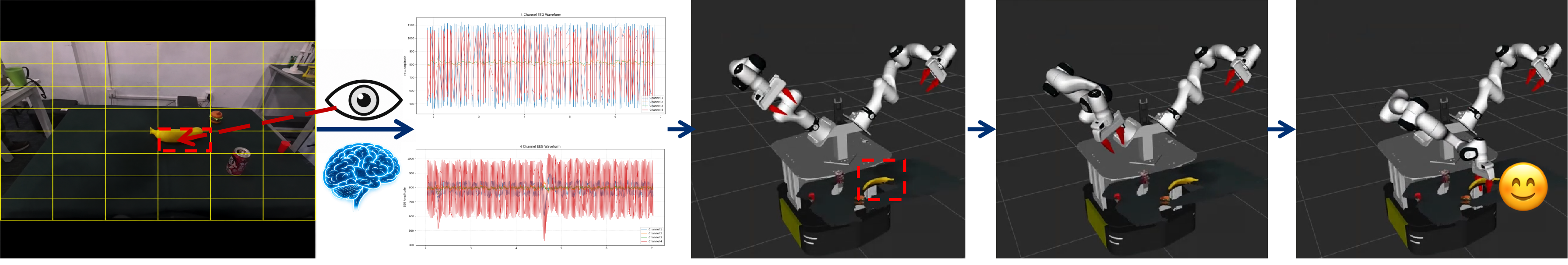}
    \caption{Qualitative visualization of offline robot target selection.
    SAM segments the four scene objects and maps their positions to candidate
    grid labels. EEG-VID supplies the 48-region EEG posterior, and the final
    target is the candidate label with the largest EEG score.}
    \label{fig:robot_visualization}
\end{figure}

EEG-VID achieves \textbf{40.24\%} window-level candidate-constrained
target-selection accuracy, exceeding the \textbf{25\%} random chance level
for four candidate objects by 15.24 percentage points. Because SAM only
defines the four eligible candidate locations and does not infer the intended
target, the above-chance result indicates that the EEG-derived spatial
posterior retains useful target information after scene-based restriction.

This result should not be directly compared with the 48-way VIG-48 Top-1
accuracy because the two evaluations use different decision spaces. The
robot-scene experiment tests whether a calibrated 48-region EEG posterior can
support object selection after being restricted to four scene-consistent
candidates. It therefore evaluates the complementary roles of visual scene
constraints and EEG intention evidence rather than unconstrained 48-way
decoding.

This experiment evaluates target selection before motion planning and does
not measure grasp success. A substantial gap remains to reliable closed-loop
assistive control. Fig.~\ref{fig:robot_visualization} illustrates the
candidate-restriction process.


\section{Discussion}

\subsection{Main Findings}
The main finding is that Stage~1 improves matched decoders across datasets and
evaluation protocols, with the clearest gains under subject shift. The
subject-wise results further show that these gains are broadly distributed
rather than driven by a small subset of participants, supporting transfer
beyond subject-overlapping cross-session evaluation.

The ablations separate the contribution of the training objective from
predictor design. Latent prediction alone improves over direct supervision,
and weak task guidance further strengthens the learned representation. The
benefit also persists with generic temporal predictors, indicating that the
transfer effect is not tied to a single predictor architecture.

Overall, the results identify task-guided latent prediction as the primary
transferable mechanism in the tested setting, while predictor design provides
an additional architecture-dependent gain. The selected task-guidance weight and predictor configuration work well in the tested settings but are not necessarily optimal for other datasets or models.

\subsection{Physiological Interpretation and Assistive Relevance}
\label{sec:origin}
We interpret VIG-48 at the wearable-device level rather than as
source-specific cortical decoding. Natural gaze orienting is permitted,
AF7/AF8 are close to the eyes, and eye movements can contribute strongly to
scalp EEG during free viewing
\cite{plochl2012combining,dimigen2020optimizing,lin2023neural,afonso2025eeg}.
Without reference EOG or eye tracking, cortical, visually evoked, and ocular contributions cannot be separated. The electrode-subset analysis is therefore descriptive and is not used for source-localization claims. The transferability of Stage~1 is assessed independently on the motor-imagery IV-2a/IV-2b benchmarks.

As a stand-alone 48-way command channel, 6.52\% Top-1 accuracy is insufficient
for practical control. Restricting the decision to four SAM-derived scene
candidates yields 40.24\% window-level target-selection accuracy versus a
25\% chance level. This supports an offline, calibrated proof of concept, not
closed-loop or clinical readiness.

\subsection{Limitations and Future Work}
Several limitations define the scope of the current study. VIG-48 contains one participant, and the robot-scene study uses subject-specific calibration. Future work will expand the visual-intention dataset to multiple participants and evaluate cross-subject transfer with less subject-specific calibration. The four-channel montage and unconstrained gaze do not allow cortical and ocular contributions to be separated. Future studies will add EOG or eye tracking and denser EEG to better distinguish these sources. Public-benchmark results use 1-s window-level
decisions, so trial-level aggregation should be evaluated for direct
comparison with standard protocols. Finally, Stage~1 requires additional computation compared with direct supervision. Future work will use more closely compute-matched comparisons and closed-loop robot experiments with online spatial-belief updating and EEG-based correction.

Accordingly, the current results support a transferable pretraining strategy
and an offline scene-informed target-selection proof of concept rather than a
deployable BCI or a source-specific neural mechanism.

\section{Conclusion}
EEG-VID combines task-guided latent predictive pretraining with supervised
EEG decoding. Stage~1 improves matched decoders under cross-session and
cross-subject shifts and remains effective when the proposed predictor is
replaced by generic recurrent or Transformer alternatives. In the robot-scene
study, restricting the EEG-derived posterior over grid regions to
scene-derived candidates yields above-chance window-level target selection
after subject-specific calibration. These results support task-guided latent
prediction as a transferable EEG pretraining strategy and motivate future
closed-loop evaluation.

\bibliographystyle{IEEEtran}
\bibliography{references}

\end{document}